\documentclass[journal]{IEEEtran}
\usepackage{cite}
\usepackage{amsmath,amssymb,amsfonts}
\usepackage{algorithmic}
\usepackage{graphicx}
\usepackage{tikz} 
\usepackage{textcomp}
\usepackage{multirow}
\usepackage[caption=false,font=footnotesize]{subfig}
\usepackage{cleveref} 
\usepackage{url}
\usepackage{threeparttable}

\def\BibTeX{{\rm B\kern-.05em{\sc i\kern-.025em b}\kern-.08em
    T\kern-.1667em\lower.7ex\hbox{E}\kern-.125emX}}
    
\newcommand{\meanstd}[2]{$#1\!\pm\!#2$}

\begin{document}
\title{Beyond Gait: Person Identification from Millimeter-Wave Point Clouds Across Activities of Daily Living}

\author{Xilai~Wang$^{*}$,
        Zixiong~Han$^{*}$,~\IEEEmembership{Graduate Student Member,~IEEE,}
        Saad~Rhanmouni,\\
        Chenzhe~Zhao,
        Yunze~Lu,
        and~Miodrag~Bolic,~\IEEEmembership{Senior Member,~IEEE}%
\thanks{$^{*}$Xilai Wang and Zixiong Han contributed equally to this work.}%
\thanks{All authors are with the Computational Analysis and Applications Research Group, School of Electrical Engineering and Computer Science, University of Ottawa, Ottawa, ON K1N 6N5, Canada (e-mail: \{xwang736, zhan045, srhan056, czhao9, ylu239, Miodrag.Bolic\}@uottawa.ca).}%
}
        
\maketitle

\begin{abstract}

Person identification from millimeter-wave (mmWave) radar point clouds has primarily focused on gait. However, since indoor walking is typically brief and fragmented, gait-based identification is largely constrained to designated passages that support continuous walking. Meanwhile, non-gait activities of daily living (ADLs) are pervasive and also encode discriminative personal signatures. In this study, we evaluate the feasibility of ADL-based identification from sparse mmWave point clouds using mm-ADL, a dataset comprising seven core ADLs collected from 11 subjects under controlled protocols. Moving beyond uniform gait introduces significant activity-dependent variations that confound identity representations. To address this challenge, we systematically benchmark three activity–identity learning paradigms: monolithic learning, joint multi-task learning, and two-stage activity-conditioned routing, under both closed-set identification (ID) and subject-disjoint re-identification (ReID) biometric settings. In parallel, we propose Dual-Stream Static-Dynamic PointNet (DS-SDPNet), an architecture tailored to radar point clouds that captures complementary spatial morphology and motion dynamics. Across seven ADLs, DS-SDPNet achieves $70.2\%$ mean closed-set identification accuracy and outperforms existing baselines on both mm-ADL and mmGait. Moreover, benchmarking the three paradigms without ground-truth activity labels during inference yields up to $68.0\%$ closed-set accuracy and $75.4\%$ ReID mAP, providing a systematic baseline and confirming the feasibility of point-cloud-based biometric identification across diverse daily activities.

\end{abstract}


\section{Introduction}\label{sec:introduction}
Human activity patterns inherently encode person-specific biometric signatures. Millimeter-wave (mmWave) point clouds provide an attractive sensing modality for activity-pattern-based identification. Compared with camera-based solutions, mmWave sensing operates reliably across varying illumination conditions and can penetrate nonconductive occlusions, enabling sensing under visual occlusion. It also preserves privacy by omitting facial images or detailed photographic appearance. A mmWave point-cloud clip contains spatial and temporal information related to a person's body geometry, posture, and movement. It provides the basis for investigating identity information in activities.

Gait is the activity pattern that has been studied most in mmWave point-cloud identification. However, it requires a sufficiently informative walking sequence. In a home or office setting, a person may walk only briefly between locations, with frequent starts, stops, turns, and halts. Therefore, gait-based identification is mainly applied in passages, corridors, aisles, or open walkways. Meanwhile, other activities of daily living (ADLs), such as sitting down, lying down, and getting up, may also contain person-specific information. Extracting identity information from these activities can extend the range of activities used for identification and broaden the possible application settings beyond passageways. Although several public mmWave point-cloud datasets have been developed for gait recognition~\cite{meng2020gait,li2022gait,mazzieri2025open,huang2026beyond}, identification across heterogeneous daily activities using mmWave point clouds remains less explored.

The extension from relatively uniform gait biometrics to diverse activity biometrics also introduces a modeling challenge: identity cues are intertwined with activity-dependent variations, making biometric feature learning more complex. Different windows from a gait point-cloud clip commonly contain the same type of motion, while an ADL point-cloud recording contains heterogeneous states and transitions, including standing/walking, sitting, lying down, and transitions among them. Activity changes both the posture and the observed body configuration.

Recent activity biometric studies have recognized this challenge~\cite{lee2017implicit,zeng2017wearia,xu2024gestureprint} and proposed different approaches. Existing strategies include learning activity and identity through a shared representation and learning activity or context separately before selecting a corresponding identity model. However, the relative effectiveness of these strategies varies with the sensing modality and has not been systematically examined for heterogeneous mmWave point-cloud ADLs. Meanwhile, radar point clouds are inherently sparse and require a tailored representation-learning model that can capture both spatial point distributions and temporal motion characteristics.

To address these challenges, this paper investigates the feasibility of mmWave point-cloud person identification beyond gait, spanning seven diverse domestic ADLs. Using our dedicated 11-subject mm-ADL dataset, we systematically benchmark three primary activity–identity learning paradigms, including monolithic learning, joint multi-task learning, and two-stage activity-conditioned routing, under both closed-set identification (ID) and subject-disjoint re-identification (ReID) protocols. In parallel, we introduce Dual-Stream Static-Dynamic PointNet (DS-SDPNet), an architecture tailored to sparse radar point-cloud sequences that couples frame-level spatial aggregation with temporal motion encoding and outperforms current baselines on both mm-ADL and the public mmGait dataset. In the feasibility validation study, DS-SDPNet achieves a $70.2\%$ mean closed-set identification accuracy across the seven ADLs under five-fold cross-validation. In the learning-paradigm benchmarking, our framework achieves up to $68.0\%$ closed-set accuracy and $75.4\%$ ReID mAP when evaluated without test-time activity labels. Both results demonstrate the viability of ADL-based biometrics. We also disentangle the distinct benefits of gallery search-space partitioning and activity-specific representation learning.


The main contributions of this paper are summarized as follows:
\begin{itemize}

\item We establish the feasibility of and a benchmark for person identification across seven diverse, non-gait ADLs using mmWave point clouds. 



\item We conduct a systematic evaluation of activity–identity paradigms on mmWave point clouds, comparing monolithic learning, joint multi-task learning, and two-stage activity-conditioned routing. 



\item We design DS-SDPNet, a state-of-the-art identity backbone tailored to sparse radar point-cloud sequences by integrating clip-level static spatial aggregation with dynamic temporal motion encoding. 

\end{itemize}

In the remainder of this paper, Section~\ref{sec:literature} reviews related work; Section~\ref{sec:mmWave} outlines the fundamentals of mmWave point clouds and the activities-of-daily-living taxonomy; Section~\ref{sec:methodology} details activity--identity modeling for point-cloud clips; Section~\ref{sec:experiments} describes the experimental setup; and Sections~\ref{sec:results_id} and~\ref{sec:results_reid} present the results and analysis for closed-set ID and subject-disjoint ReID. Section~\ref{sec:backbone_analysis} analyzes point-cloud representation-learning backbones, Section~\ref{sec:limitations} discusses study limitations, and Section~\ref{sec:conclusion} concludes the paper.

\section{Related Work}\label{sec:literature}
\subsection{Activity-Based Person Identification}
Activity motion pattern-based person identification has been studied with various sensing modalities. Iosifidis et al.~\cite{iosifidis2011activity} use multi-camera body masks for identification from activities including walking, running, jumping, and waving. Azad et al.~\cite{azad2024activity} use RGB videos, and Wang et al.~\cite{wang2018learning} use 3-D skeletons extracted from RGBD data for person identification from daily activities. Daily activity patterns have also been captured using RFID~\cite{huang2019id} and wearable inertial sensors~\cite{chen2020metier,shin2023identifying}. These studies support investigating both body configuration and the personal manner of performing an activity as sources of identity information.

\subsection{mmWave Radar-Based Person Identification}
Radar sensing has also been used for biometric studies. The person motion pattern is mainly extracted from micro-Doppler signatures and point clouds. Micro-Doppler signatures have been used to identify people from specific motions such as jumping, crawling, walking, and boxing~\cite{yang2019person,lang2020person}, from sit-to-stand and stand-to-sit movements~\cite{saho2020accurate,li2025sit}, and specific gestures~\cite{wu2025joint}. Compared with micro-Doppler signatures, mmWave point clouds capture aspects of a subject's body configuration and can also support multi-person tracking. mmWave point clouds have been used for gesture-based identification~\cite{xu2024gestureprint} and gait-based identification~\cite{meng2020gait,cheng2021person,wang2022stpointgcn,huang2023hdnet,xue2023fine,wang2025open}. Among them, gait-based identification has been studied in different application scenarios: Meng et al.~\cite{meng2020gait} study co-existing people, Cheng and Liu~\cite{cheng2021person} study person ReID, and Wang et al.~\cite{wang2025open} study open-set identification under occlusion. Subsequent work has proposed different spatio-temporal modeling architectures to improve identification accuracy~\cite{wang2022stpointgcn,huang2023hdnet,xue2023fine}. As radar point cloud has also been used for human activity recognition across a broad range of activities \cite{wang2021m, yu2022noninvasive, yang2023mm, cui2024milipoint}, we are motivated to extend mmWave point-cloud identification from gait and gestures to ADLs for a broader application setting.

\subsection{Integrating Activity and Identity Information}
Activity biometrics faces the challenge that identity cues and activity-related features are intertwined. Existing activity biometric studies address this challenge through three paradigms: monolithic learning, joint/multi-task representation learning and activity/context-conditioned identification. Monolithic learning trains a single model across all activity data, relying entirely on the backbone's representative capacity to map diverse motion patterns to individual identities \cite{xu2024gestureprint}. Joint and multi-task approaches optimize activity and identity recognition together through parameter sharing or cross-task feature exchange~\cite{wang2018learning,chen2020metier,azad2024activity,wu2025joint}. Such joint learning can also incorporate explicit feature disentanglement, as in language-guided separation of biometric and motion features~\cite{azad2025disenq} and contrastive decoupling of identity and activity~\cite{wang2026wifi}. Activity/context-conditioned approaches learn activity/context and identity in separate stages, first inferring the activity/context and then applying a corresponding identification or authentication model~\cite{lee2017implicit,zeng2017wearia,xu2024gestureprint}.

\section{mmWave Point Clouds and ADL Dataset}\label{sec:mmWave}
\subsection{Radar Sensing and Point-Cloud Generation}
The mmWave radar used in our study is a frequency-modulated continuous-wave (FMCW) radar. It transmits linear frequency-modulated chirps. A chirp can be described as 
\begin{equation}
\begin{aligned}
s(t)&=\exp\!\left\{j2\pi\left(f_ct+\frac{S}{2}t^2\right)\right\},\\
0&\leq t\leq T_c .
\end{aligned}
\label{eq:chirp}
\end{equation}
where $f_c$ is the starting frequency, $B_w$ is the bandwidth, $T_c$ is the chirp duration, and $S=B_w/T_c$ is the chirp slope. The received signal is dechirped to produce a beat signal, which is then processed using a fast-time Fourier transform (FFT) to estimate target range. A slow-time Doppler FFT across chirps estimates radial velocity, while phase differences across the virtual antenna array are used to estimate azimuth and elevation angles. The resulting range resolution is $\Delta r=c/(2B_w)$. The radar configuration is summarized in Table~\ref{tab:radar_params}.

\begin{table}[h]
\caption{Radar waveform configuration.}
\label{tab:radar_params}
\centering
\setlength{\tabcolsep}{4pt}
\begin{tabular}{lclc}
\hline
Parameter & Value & Performance & Value \\
\hline
$f_{\min}$ & 60.75 GHz & $\Delta R$ & 0.084 m \\
$f_{\max}$ & 63.98 GHz & $R_{\max}$ & 8.09 m \\
$B_w$ & 1780 MHz & $\Delta v$ & 0.093 m/s \\
$T_c$ & 89.10 $\mu$s & $v_{\max}$ & 4.50 m/s \\
$N_s$ & 96 & $\Delta\varphi,\Delta\theta$ & $28.6^\circ$ \\
$N_{\mathrm{chirp}}$ & $3\times96$ & FoV & $\pm60^\circ$ \\
$T_{\mathrm{frame}}$ & 55 ms & Frame rate & 18.2 fps \\
\hline
\end{tabular}
\end{table}

A range FFT is followed by mean subtraction across chirps to suppress static clutter. Capon beamforming produces a range--azimuth heatmap, and constant false-alarm rate detection extracts candidate cells. Elevation and Doppler are estimated at the detected cells, producing points with $[x,y,z,v_r,\mathrm{SNR}]$. A group tracker associates points across frames and removes unassociated detections, leaving the target point cloud used in this work.

In this processing chain, static clutter removal suppresses the subject’s radar returns when the subject becomes fully stationary. Consequently, the near-static activities (e.g. sitting and lying down) are observable mainly through residual movement before a stationary state is reached.

\subsection{ADL Taxonomy}\label{sec:ADL}
\begin{figure}[h]
    \centering
    \includegraphics[width=\columnwidth]{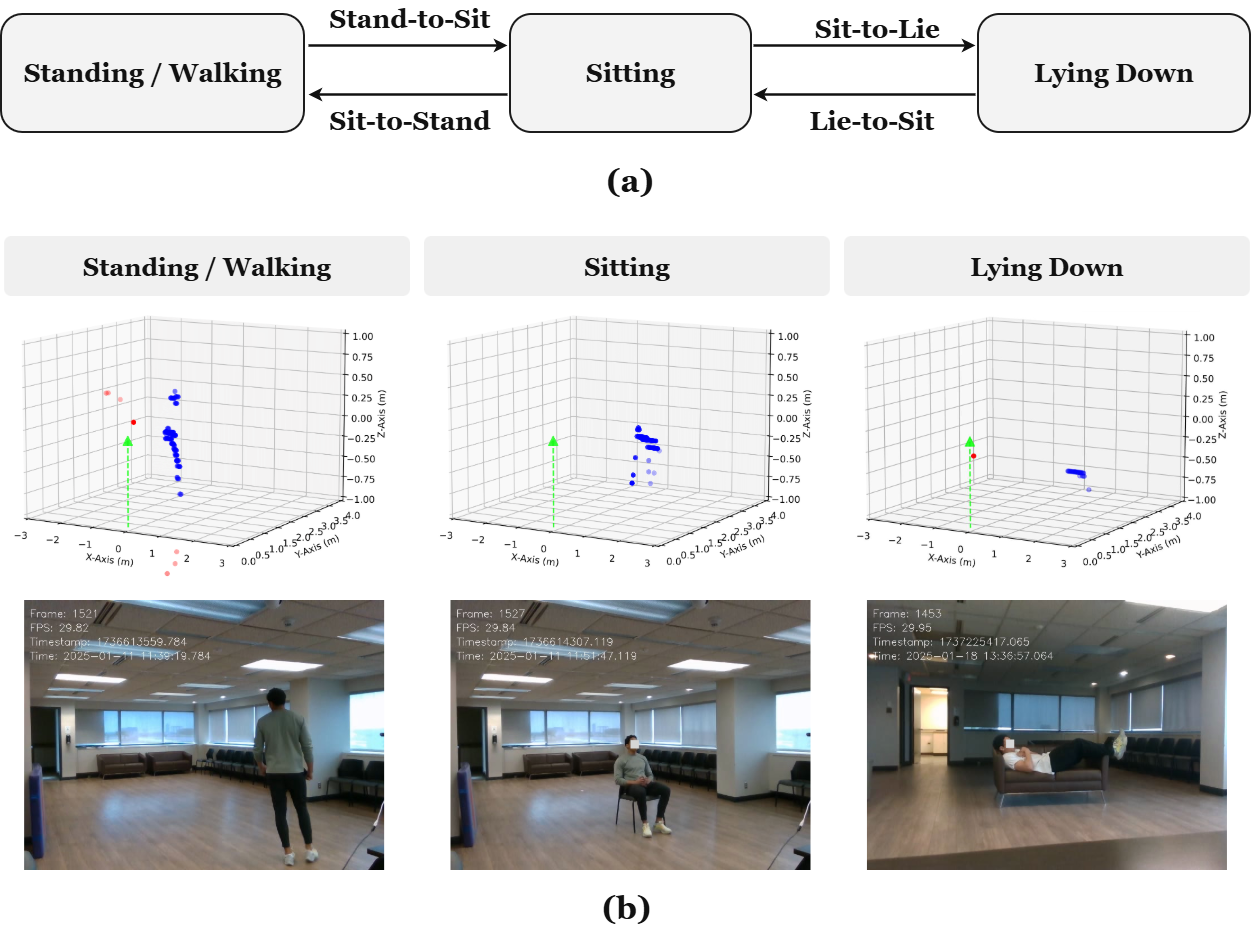}
    \caption{ADL taxonomy and representative observations in mm-ADL. (a) State-transition relationships among the three major states and four transition activities. (b) Representative radar point clouds (top) and corresponding recorded images (bottom) for Standing / Walking, Sitting, and Lying Down.}
    \label{fig:adl_data}
\end{figure}

The activities of daily living (ADLs) considered in our study follow a state-and-transition-based taxonomy that describes consecutive indoor movements, consistent with transition-aware HAR~\cite{reyes2016transition,amin2019radar}. The seven-activity ethogram contains three states (Standing/Walking, Sitting, and Lying Down) and four transitions (Stand-to-Sit, Sit-to-Lie, Lie-to-Sit, and Sit-to-Stand). These activities are connected through the transitions shown in Fig.~\ref{fig:adl_data} (a).

Standing and walking are grouped into a single activity class because limited indoor space often does not allow a long continuous walk, resulting in recorded motion with frequent starts and stops. Sitting and lying down are referred to as near-static states in this work. Point cloud examples of the state activities are shown in Fig.~\ref{fig:adl_data} (b).


\subsection{mm-ADL Data Collection}

We use a Texas Instruments IWR6843ISK-ODS radar to collect
mmWave point clouds. It is equipped with a multiple-input
multiple-output (MIMO) antenna array with a
$120^\circ$ field of view in both azimuth and elevation,
providing wide angular coverage for human motion sensing.
The radar is mounted at a height of approximately 1.5~m,
and the monitored range is 0.9--7~m. An Intel RealSense RGB
camera records synchronized video for annotating subject
identity, activity, and activity start and end times.

Eleven subjects, including six males and five females, with heights ranging from 1.60 to 1.85~m, were recruited as volunteers for this study. Each subject performed multiple predefined activity sequences in a $4\times8$ m conference room. We collected activity sequences with three protocols:
\begin{enumerate}
    \item Walk protocol: Random wandering with random turns and halts.
    \item Sit protocol: Walk to a chair, sit down, remain seated for a short interval, stand up, and return.
    \item Lie protocol: Walk to a bed/sofa, lie down, remain lying for a short interval, get up, and return.
\end{enumerate}
Each subject repeated each sequence 50 times. All subjects followed the same execution protocol under a fixed radar geometry, so the evaluation focuses on differences in subject build and movement patterns under the same viewpoint and environment. Data were collected on multiple days, and subjects changed outerwear to reduce dependence on clothing material and style. The study was approved by the University of Ottawa Research Ethics Boards (REB), ethics ID: H-11-25-12304.

Table~\ref{tab:activity_avg_frames} reports the average number
of frames per activity occurrence. Transitions are relatively
short, whereas Standing/Walking is substantially longer.
The observed Sitting and Lying Down intervals are also
limited because static clutter removal suppresses stationary
subject reflections. Consequently, the point clouds available
during these intervals primarily reflect residual postural
motion.

\begin{table}[h]
\centering
\caption{Average frames per activity type in mm-ADL.}
\label{tab:activity_avg_frames}
\setlength{\tabcolsep}{12pt}
\begin{tabular}{lc}
\hline
Activity & Average frames \\
\hline
Standing/Walking & 161.03 \\
Sitting & 31.02 \\
Lying Down & 29.98 \\
Lie-to-Sit & 27.24 \\
Sit-to-Lie & 24.81 \\
Stand-to-Sit & 14.73 \\
Sit-to-Stand & 11.96 \\
\hline
\end{tabular}
\end{table}

\begin{figure*}[!t]
\centering
\includegraphics[width=\textwidth]{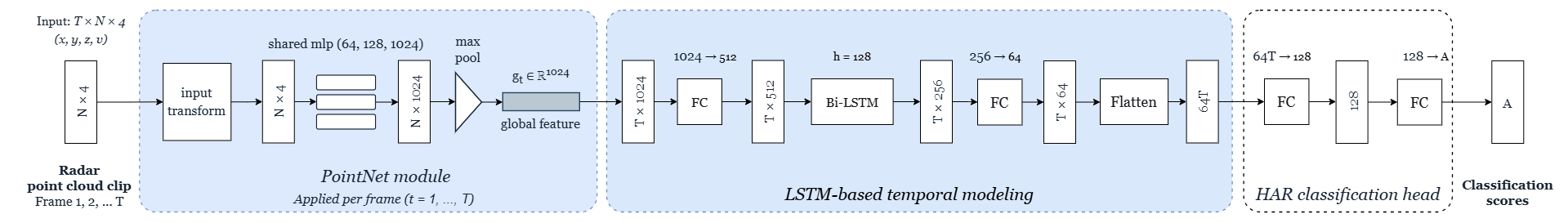}
\caption{PointNet-LSTM architecture for human activity recognition.}
\label{fig:har_router_architecture}
\end{figure*}

\subsection{Point-Cloud Preprocessing and Clip Construction}

Each point cloud clip is represented as $\mathcal{P}\in\mathbb{R}^{T\times N\times C}$, where each point has $C=4$ attributes $(x,y,z,v_r)$. To reduce dependence on the person’s absolute horizontal location, we center each frame in the horizontal plane. For frame $t$, the centroid and normalized horizontal coordinates are computed as follows:
\begin{equation}
\begin{aligned}
\mathbf{c}_t&=\frac{1}{N}\sum_{n=1}^{N}(x_{t,n},y_{t,n}),\\
(\tilde{x}_{t,n},\tilde{y}_{t,n})&=(x_{t,n},y_{t,n})-\mathbf{c}_t .
\end{aligned}
\label{eq:normalisation}
\end{equation}
The height $z_{t,n}$ remains unchanged because it is measured relative to the ground and provides a cue about body geometry. 


We segment the recordings using a fixed-length sliding
window of $T=20$ frames. Each window is assigned the label using frame-wise majority voting. Each frame inherits a label from the synchronized video timestamp. The window is assigned the label that appears most frequently ($>50\%$) within those $T$ frames \cite{reyes2016transition}. The window length is selected with reference
to the shortest class-averaged activity duration. A window containing a complete activity
of approximately this duration would therefore contain
more corresponding activity frames than frames from surrounding
activities.

Each frame is
represented by 128 points: frames containing more points
are randomly downsampled, while frames containing fewer
points are padded by sampling the available points with
replacement \cite{meng2020gait}. To balance the activity classes, we retain
50 clips per activity per subject, each sampled from a
distinct occurrence of that activity. When multiple sliding
windows cover the same activity occurrence, only one clip
is retained. The resulting mm-ADL dataset contains
3,850 clips of 20 frames each, totaling 77,000 frames across
the selected clips. Each clip is represented as
$\mathcal{P}\in\mathbb{R}^{20\times128\times4}$.
The dataset is publicly available at
\url{https://github.com/OwenHan10/mm-ADL}.

\section{Activity--Identity Modeling}
\label{sec:methodology}

This section addresses two complementary challenges in mmWave biometric sensing: extracting discriminative spatiotemporal features from sparse point clouds, and modeling intertwined activity and identity information.

\subsection{Point-Cloud Representation Learning}
\label{subsec:representation_learning}

Both activity recognition and person identification can
be formulated as point-cloud clip representation learning
tasks. Human activity recognition (HAR) aims to distinguish
different posture and motion patterns, while person
identification requires capturing finer-grained geometric
and temporal motion characteristics. As mmWave point
clouds provide limited appearance information,
inter-person differences within the same activity
can be less pronounced than differences across activities. Therefore, activity recognition task can be regarded as a foundation task and the identification task as an advanced task.

\subsubsection{PointNet-LSTM for Activity Recognition}
\label{subsec:har_router}

We start with the activity-recognition task to build the representation learner. We adopt a spatiotemporal hybrid architecture, using PointNet~\cite{qi2017pointnet} to extract frame-wise spatial features from point clouds and a Bi-LSTM to model temporal dependencies across frames. The architecture is illustrated in Fig.~\ref{fig:har_router_architecture}.

The model processes the normalized clip
$\tilde{\mathcal{P}}$ using a frame-wise PointNet encoder
with a $4\times4$ input transform and shared MLP widths
of 64, 128, and 1024. The encoder is applied to every
frame of the clip with tied weights. Batch normalization
and ReLU follow the MLP layers, and symmetric max pooling
over the $N$ points produces one 1024-D descriptor per
frame.

Each descriptor is projected to 512 dimensions using a
fully connected (FC) layer followed by ReLU activation
and dropout. A single-layer bidirectional LSTM with
128 hidden units per direction then produces a
$T\times256$ sequence. A second FC layer with ReLU
activation and dropout projects each time step from
256 to 64 dimensions. The dropout probability is 0.1
for both layers.

The $T$ projected time steps are concatenated into
a $64T$-D vector. Retaining the ordered time steps
helps distinguish opposite transitions, such as
Stand-to-Sit and Sit-to-Stand. The classification
head consists of an FC layer that reduces the
dimension from $64T$ to 128, followed by batch
normalization, ReLU activation, dropout, and a
final FC layer that produces the activity-logit
vector
$\boldsymbol{\ell}=[\ell_1,\ldots,\ell_A]
\in\mathbb{R}^{A}$,
where $\ell_a$ is the unnormalized score for
activity class $a$.

These logits are converted into the activity-probability
vector $\mathbf{p}=[p_1,\ldots,p_A]$ using
$\mathbf{p}=\operatorname{softmax}(\boldsymbol{\ell}/\tau)$.
Specifically,
\begin{equation}
p_a =
\frac{\exp(\ell_a/\tau)}
{\sum_{a'=1}^{A}\exp(\ell_{a'}/\tau)},
\qquad a=1,\ldots,A,
\label{eq:har_activity_probabilities}
\end{equation}
where $\tau=1$.

The PointNet-LSTM model is trained using cross-entropy
loss. Let $\ell_{i,a}$ denote the predicted activity
logit for class $a$ and $a_i$ the ground-truth activity
label of clip $i$. For $A$ activity classes, the HAR
objective is
\begin{equation}
\mathcal{L}_{\mathrm{HAR}}
=
-\frac{1}{B}\sum_{i=1}^{B}
\log
\frac{\exp(\ell_{i,a_i})}
{\sum_{a'=1}^{A}\exp(\ell_{i,a'})}.
\label{eq:har_loss}
\end{equation}


\begin{figure*}[!t]
\centering
\includegraphics[width=\textwidth]
{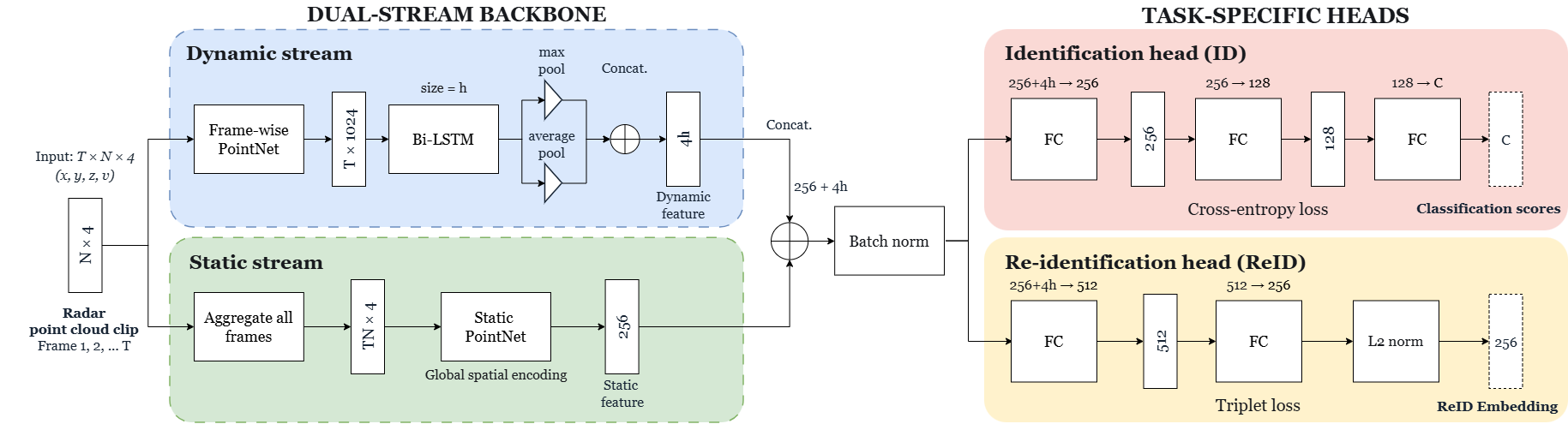}
\caption{DS-SDPNet architecture.}
\label{fig:ds_sdpnet_architecture}
\end{figure*}

\subsubsection{DS-SDPNet for Identity Representation}
\label{sec:ds_sdpnet}

The dual-stream static-dynamic PointNet (DS-SDPNet)
identification model is built upon the spatial-temporal
representation design of PointNet-LSTM. We introduce
an additional static stream to aggregate spatial
point distributions across the clip, capturing
information related to body configuration.
The original PointNet-LSTM design serves as the
basis of the dynamic stream and is further enhanced
with temporal average and max pooling over LSTM
hidden states to summarize consistently expressed
motion features and salient feature activations,
respectively.

The resulting static and dynamic representations
are fused to form a unified identity representation,
as illustrated in
Fig.~\ref{fig:ds_sdpnet_architecture}.

\paragraph{Dynamic Stream}
The dynamic stream processes the clip as an ordered
sequence of frames. It adopts the same frame-wise
PointNet encoder architecture as the PointNet-LSTM
model described above, producing a 1024-D descriptor
per frame through symmetric max pooling.

The resulting sequence is processed by a single-layer
bidirectional LSTM with hidden size $h$ per direction.
Temporal average and max pooling are applied
separately to the $T$ bidirectional outputs, and
the resulting vectors are concatenated to form a
$4h$-D dynamic feature.

\paragraph{Static Stream}
The static stream aggregates the $T$ frames into
a single point cloud containing $T\times N$ points.
A separate PointNet, equipped with a $4\times4$
input transform and shared MLP layers of widths
64, 128, and 256, produces a 256-D static feature
through global max pooling.

Unlike the dynamic stream, which explicitly models
the temporal sequence, the static stream aggregates
spatial point distributions over the entire clip.

\paragraph{Feature Fusion}
The static and dynamic features are concatenated
and batch normalized to obtain a $(256+4h)$-D fused
representation $\mathbf{z}$, which is passed to
the task-specific identity head.

We set $h=64$ for ID and $h=128$ for ReID,
yielding 512-D and 768-D fused representations,
respectively.


\subsection{Task-Specific Identity Heads}
\label{subsec:heads}

The ID and ReID tasks use the same DS-SDPNet backbone
architecture but different heads and training losses.
Each activity-specific expert has its own independently
trained identification or re-identification head.

\subsubsection{Identification Head}

Closed-set ID assigns each input clip to one of
the $C$ identities observed during training.
The identification head transforms the fused
DS-SDPNet features into identity logits through
three fully connected (FC) layers. The first two
project the features to 256 and 128 dimensions,
respectively, each followed by batch normalization,
ReLU activation, and dropout. The final layer
produces $C$ logits corresponding to the enrolled
identities.

Let $s_{i,c}$ denote the identity logit for
subject class $c$ and $y_i$ the ground-truth
identity label of clip $i$. The identification
head is optimized using cross-entropy loss over
a mini-batch of $B$ clips:
\begin{equation}
\mathcal{L}_{\mathrm{ID}}
=
-\frac{1}{B}\sum_{i=1}^{B}
\log
\frac{\exp(s_{i,y_i})}
{\sum_{c'=1}^{K}\exp(s_{i,c'})}.
\label{eq:id_loss}
\end{equation}

\begin{figure*}[!t]
    \centering
    \includegraphics[width=\textwidth]{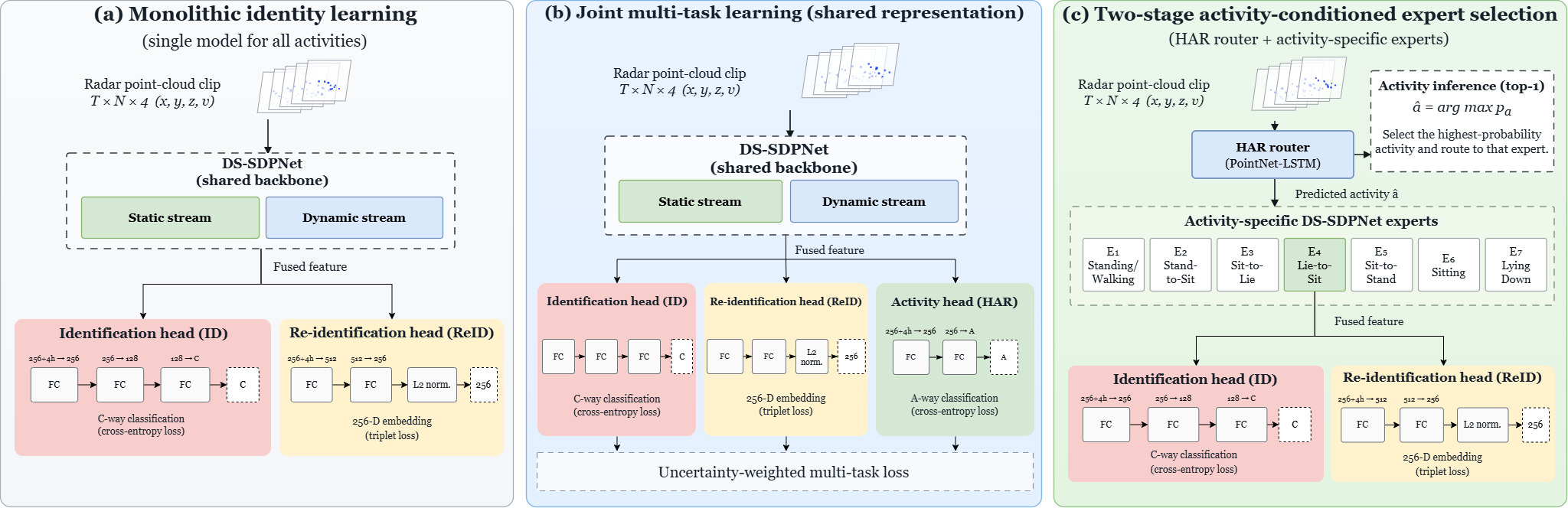}
    \caption{Three activity--identity learning strategies: (a) monolithic identity learning using a single model for all activities; (b) joint multi-task learning with a shared DS-SDPNet representation and parallel task-specific identity and HAR heads; and (c) two-stage activity-conditioned learning with a separate HAR router and hard top-1 selection among seven activity-specific experts. ID and ReID heads depict alternative task configurations.}
    \label{fig:strategies}
\end{figure*}


\subsubsection{Re-identification Head}

Open-set ReID requires identity-discriminative
embeddings that generalize to unseen subjects.
The ReID head transforms the fused DS-SDPNet
features into a 256-dimensional embedding through
two fully connected layers. The first projects
the features to 512 dimensions, followed by
batch normalization, ReLU activation, and dropout,
while the second produces a 256-dimensional
representation. The resulting feature vector
is $L_2$-normalized to obtain the identity
embedding
\begin{equation}
\label{eq:reid_embedding}
\mathbf{e} =
\frac{\psi(\mathbf{z})}
{\|\psi(\mathbf{z})\|_2}
\in \mathbb{S}^{F-1},
\end{equation}
where $\mathbf{z}$ is the $(256+4h)$-dimensional
fused DS-SDPNet feature, $\psi(\cdot)$ denotes
the two-layer projection head, and $F=256$
is the embedding dimension.

The normalized embeddings are optimized using
batch-hard triplet loss with online hard-example
mining. Let $\mathbf{e}_i$ and $\mathbf{e}_j$
denote the embeddings of an anchor clip $i$
and a candidate clip $j$ within a mini-batch,
with corresponding identity labels $y_i$
and $y_j$. Their Euclidean distance is defined as
$D(\mathbf{e}_i,\mathbf{e}_j)
=\|\mathbf{e}_i-\mathbf{e}_j\|_2$.

For each anchor $i$, batch-hard mining selects
the farthest positive (same-subject) clip
and the nearest negative (different-subject)
clip, yielding
\begin{equation}
d_i^+ =
\max_{\substack{j:y_j=y_i,\,j\ne i}}
D(\mathbf{e}_i,\mathbf{e}_j),
\end{equation}
\begin{equation}
d_i^- =
\min_{j:y_j\ne y_i}
D(\mathbf{e}_i,\mathbf{e}_j).
\end{equation}

The ReID loss is formulated as
\begin{equation}
\mathcal{L}_{\mathrm{ReID}} =
\frac{1}{B}\sum_{i=1}^{B}
[d_i^+-d_i^-+m]_+,
\label{eq:reid_loss}
\end{equation}
where $[x]_+=\max(x,0)$, $m=0.2$ is the margin,
and $B$ denotes the number of valid anchors
in the mini-batch. Anchors without a valid
positive or negative are excluded from the
loss computation.

The loss encourages same-subject embeddings
to remain closer than different-subject
embeddings by at least the specified margin,
enabling identity retrieval without a fixed
classifier over the test subjects.

\subsection{Activity--Identity Learning Strategies}
\label{subsec:learning_strategies}

We investigate three strategies for integrating activity and identity information from mmWave point-cloud clips: monolithic identity learning \cite{xu2024gestureprint}, joint multi-task learning with a shared representation~\cite{wang2018learning,chen2020metier,azad2024activity,wu2025joint}, and two-stage activity-conditioned expert selection ~\cite{lee2017implicit,zeng2017wearia,xu2024gestureprint}, as illustrated in Fig.~\ref{fig:strategies}. All three strategies use the DS-SDPNet backbone and task-specific heads detailed in Section~\ref{sec:ds_sdpnet} and~\ref{subsec:heads}, ensuring that performance comparisons isolate the structural paradigm. Closed-set ID and open-set ReID are configured as parallel, task-specific evaluation settings.




\subsubsection{Monolithic Identity Learning}
The monolithic baseline trains a single DS-SDPNet model across all activities using identity supervision alone (Fig.~\ref{fig:strategies}(a)). The shared backbone extracts an entangled $(256+4h)$-dimensional fused representation that directly feeds the task-specific identity head: either the three-layer MLP for ID optimized via $\mathcal{L}_{\text{ID}}$ (Eq.~\eqref{eq:id_loss}) or the two-layer projection head with $\ell_2$-normalization for ReID optimized via $\mathcal{L}_{\text{ReID}}$ (Eq.~\eqref{eq:reid_loss}). This model must accommodate the wide variance across distinct postures and dynamic patterns within a single parameter set without explicit activity conditioning.





\subsubsection{Joint Multi-Task Learning}
The joint multi-task learning (MTL) framework trains a single shared DS-SDPNet backbone simultaneously on identity and activity recognition (Fig.~\ref{fig:strategies}(b)). The fused representation $z \in \mathbb{R}^{256+4h}$ is shared across two parallel heads: the corresponding identity head, ID or ReID, and an auxiliary HAR head.

The auxiliary HAR head comprises two fully connected layers. The first layer projects the $(256+4h)$-dimensional fused feature to 256 dimensions, followed by batch normalization, ReLU activation, and a dropout probability of $0.1$. The second linear layer projects the 256-D feature to $A$ activity logits. The head is supervised using the cross-entropy loss $\mathcal{L}_{\text{HAR}}$ formulated in Eq.~\eqref{eq:har_loss}. The ID and ReID heads are identical to the monolithic identity learning model.

To balance the identity objective ($\mathcal{L}_{\text{task}} \in \{\mathcal{L}_{\text{ID}}, \mathcal{L}_{\text{ReID}}\}$) and the auxiliary activity objective ($\mathcal{L}_{\text{HAR}}$) without manually tuned hyperparameters, we implement homoscedastic uncertainty weighting~\cite{kendall2018multi}. Each task $i \in \{\text{task}, \text{HAR}\}$ is parameterized by a learnable log-variance $s_i \equiv \log \sigma_i^2$, yielding the overall joint multi-task objective:
\begin{equation}
    \mathcal{L}_{\text{MTL}} = \frac{1}{2\sigma_{\text{task}}^2}\mathcal{L}_{\text{task}} + \frac{1}{2\sigma_{\text{HAR}}^2}\mathcal{L}_{\text{HAR}} + \log(\sigma_{\text{task}}\sigma_{\text{HAR}}),
    \label{eq:uncertainty_loss}
\end{equation}
which is implemented in log-variance space as:
\begin{equation}
    \mathcal{L}_{\text{MTL}} = \sum_{i \in \{\text{task}, \text{HAR}\}} \left( \exp(-s_i)\mathcal{L}_i + \frac{1}{2}s_i \right).
    \label{eq:uncertainty_logvar}
\end{equation}
The log-variance parameters $s_i$ are initialized to zero and updated via backpropagation alongside network parameters, enabling data-driven loss balancing.







\subsubsection{Two-Stage Activity-Conditioned Learning}
The two-stage framework decouples activity recognition and identity learning into specialized, modular stages (Fig.~\ref{fig:strategies}(c)). During training, a standalone PointNet-LSTM model is trained as an upfront HAR router using Eq.~\eqref{eq:har_loss}, while $A$ independent DS-SDPNet experts $\{E_1, \dots, E_A\}$ are trained in parallel, each specialized exclusively on data belonging to a single activity. Each expert possesses an identical DS-SDPNet backbone and task head (ID or ReID) containing its own independent set of weights.

At inference time, an input point-cloud clip $\mathcal{P}$ is first evaluated by the HAR router to infer the activity class:
\begin{equation}
    \hat{a} = \arg\max_{a \in \{1,\dots,A\}} p_a.
    \label{eq:router_decision}
\end{equation}
The clip is subsequently dispatched exclusively to expert $E_{\hat{a}}$, whose backbone generates an activity-conditioned $(256+4h)$-dimensional fused representation. For closed-set ID, $E_{\hat{a}}$ computes identity logits over the enrolled subjects; for open-set ReID, $E_{\hat{a}}$ produces a 256-D $\ell_2$-normalized embedding that is ranked against gallery embeddings populated by the same expert.

We also evaluate oracle routing, in which the ground-truth activity selects the expert, to assess the effect of activity-prediction errors.

\section{Experimental Setup}\label{sec:experiments}

\subsection{Closed-Set Identification Protocol}
For closed-set ID, all 11 subjects appear in training and testing. For each subject and activity, the 50 non-overlapping clips are partitioned into five contiguous blocks of 10. Fold $k$ used the $k$th block for testing and the remaining 40 clips for training, yielding 3,080 training and 770 test clips per fold. The HAR router and identification experts are trained separately within each fold, with test clips evaluated regardless of routing correctness.

\subsection{Subject-Disjoint ReID Protocol} 
For subject-disjoint ReID, training and test identities are disjoint. We use six folds, each comprising nine training subjects and two held-out test subjects, arranged so that every subject appears in a test fold at least once. The HAR router and ReID experts are trained exclusively on the training subjects.

\begin{figure*}[!t]
\centering
\includegraphics[width=0.97\textwidth]{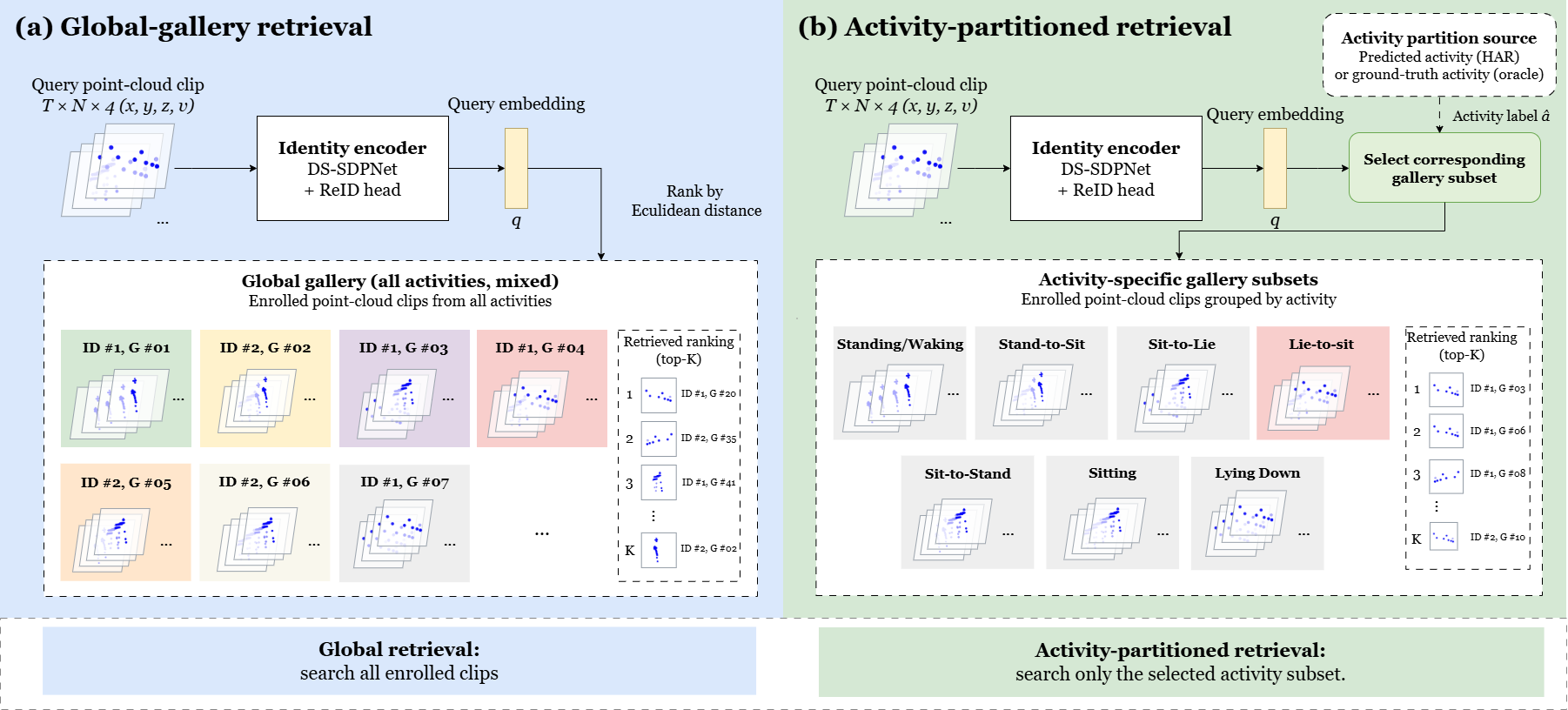}
\caption{ReID gallery evaluation protocols. (a) Global-gallery retrieval ranks enrolled clips from all activity types. (b) Activity-partitioned retrieval restricts matching to the selected activity subset, determined by predicted activity or the ground-truth activity in the oracle reference. In both protocols, gallery clips are ranked by Euclidean distance to the query embedding; multiple clips may share the same subject identity. In the retrieved ranking, ID denotes the subject identity, and G denotes the clip index in the gallery. Gallery clips are ranked by their Euclidean distances to the query embedding.}
\label{fig:gallery_protocols}
\end{figure*}

Evaluation follows an enrollment-and-retrieval protocol. For each fold, clips from the held-out subjects are grouped by activity predicted by the trained HAR router. Within each subject--predicted-activity group, the first 20\% of clips form the query set, while the remaining clips form the enrolled gallery. The two-stage model compares each query against the corresponding activity-specific gallery using its selected ReID expert, without updating model weights. Gallery identities are provided during enrollment, while query identities are determined through embedding-based retrieval. Since the gallery composition directly influences the ReID performance, we implement both global-gallery and activity-partitioned gallery for comparisons, as shown in Fig.~\ref{fig:gallery_protocols}.

\subsection{ReID Gallery Construction and Comparison Protocols}
\subsubsection{Global-Gallery Retrieval}
The global gallery includes enrolled clips from all activities. For monolithic and joint multi-task models, each query is encoded using the corresponding shared identity backbone and compared with the global gallery without requiring an activity label. Gallery entries are ranked by increasing Euclidean distance to the query embedding.

\subsubsection{Activity-Partitioned Retrieval}
For two-stage ReID, labeled enrollment clips are grouped by activity and encoded using the corresponding activity-specific expert. The HAR router selects the activity of an incoming query, determining both the expert used to encode it and the gallery subset searched for matching. Oracle partitioning uses the ground-truth activity label as a reference. The two-stage retrieval procedure thus combines activity-dependent representation learning with activity-dependent gallery selection.

\subsubsection{Controlled Activity-Partitioned Comparison}
To examine the influence of gallery construction separately from identity representation learning, we additionally evaluate the monolithic and joint multi-task embeddings with activity-partitioned galleries. The activity partition is supplied externally by the HAR router for this controlled comparison; it is not an intrinsic requirement of either shared-model approach. The activity-specific expert bank is evaluated using the corresponding activity partition. The native and controlled settings are reported separately.


\subsection{Baselines and Implementation Details}

All models are trained from scratch using the Adam optimizer
with a weight decay of $10^{-3}$, a dropout rate of $0.1$,
and a maximum of 500 epochs in the reported main configurations.
The ID and ReID experts use learning rates of
$10^{-3}$ and $10^{-4}$, with batch sizes of 25 and 60,
respectively.
The HAR router is trained with a learning rate of $10^{-3}$
and a batch size of 25.
All experiments use a fixed random seed of 42.

\subsection{Evaluation Metrics}
ID and HAR performance are evaluated using classification accuracy, defined as the proportion of test clips correctly assigned to their ground-truth identity or activity class. Results are averaged across the corresponding cross-validation folds.

For ReID, each query embedding is compared against all gallery
embeddings using Euclidean distance, and gallery samples are
ranked in ascending order of distance. Let $\mathcal{Q}$
denote the query set, $\mathcal{R}_q$ the set of correct gallery
matches for query $q$, and $r_q^{\min}$ the rank of the
highest-ranked correct match. The cumulative matching
characteristic (CMC) at rank $k$ is defined as
\begin{equation}
\mathrm{CMC}@k =
\frac{1}{|\mathcal{Q}|}
\sum_{q\in\mathcal{Q}}
\mathbb{I}(r_q^{\min}\leq k),
\label{eq:cmc}
\end{equation}
where $\mathbb{I}(\cdot)$ is the indicator function.
We report Rank-1 accuracy ($\mathrm{CMC}@1$).

While CMC considers only the highest-ranked correct match,
mean average precision (mAP) evaluates the overall retrieval
ranking. Let $r_{q,j}$ denote the rank of the $j$th correct
match for query $q$, with correct matches ordered by rank.
The average precision (AP) for query $q$ and mAP are
defined as
\begin{equation}
\mathrm{AP}_q =
\frac{1}{|\mathcal{R}_q|}
\sum_{j=1}^{|\mathcal{R}_q|}
\frac{j}{r_{q,j}},
\label{eq:ap}
\end{equation}
\begin{equation}
\mathrm{mAP} =
\frac{1}{|\mathcal{Q}|}
\sum_{q\in\mathcal{Q}}\mathrm{AP}_q.
\label{eq:map}
\end{equation}
Rank-1 measures the proportion of queries whose top-ranked
gallery sample has the correct identity, whereas mAP
evaluates the ranking of all relevant gallery samples.
ReID results are averaged across the corresponding
cross-validation folds.

\section{Closed-Set Identification Results}\label{sec:results_id}

\subsection{Identification Feasibility Across ADLs} \label{subsec:adl_feasibility}

We first investigate the feasibility of mmWave point-cloud identification across diverse ADLs under known activity class (oracle routing). Table~\ref{tab:per_activity_accuracy_comparison} reports the five-fold cross-validation results across seven activities for all evaluated backbones. Across all four models, every activity achieves an identification accuracy far exceeding the balanced chance level of $1/11 \approx 9.1\%$, with the lowest baseline floor at $48.4\%$ and top models exceeding $62.5\%$, confirming that identity-discriminative spatial and motion cues are consistently extractable throughout daily activities.

\begin{table}[h]
\centering
\caption{Per-activity ID accuracy under oracle routing on mm-ADL (\%).}
\label{tab:per_activity_accuracy_comparison}
\renewcommand{\arraystretch}{1.08}
\setlength{\tabcolsep}{3.3pt}
\begin{tabular}{lcccc}
\hline
Activity & DS-SDPNet & HDNet & DSFE+LGTE & SRPNet \\
\hline
Sit-to-Lie &
\meanstd{\mathbf{80.8}}{4.2} &
\meanstd{80.2}{4.0} &
\meanstd{79.1}{4.8} &
\meanstd{72.4}{2.9} \\

Lie-to-Sit &
\meanstd{\mathbf{74.6}}{2.0} &
\meanstd{72.9}{0.8} &
\meanstd{71.7}{2.2} &
\meanstd{64.1}{3.6} \\

Sit-to-Stand &
\meanstd{\mathbf{71.7}}{2.7} &
\meanstd{65.2}{1.8} &
\meanstd{62.9}{3.2} &
\meanstd{48.4}{3.1} \\

Stand-to-Sit &
\meanstd{\mathbf{69.8}}{2.3} &
\meanstd{65.3}{2.3} &
\meanstd{64.8}{2.1} &
\meanstd{51.2}{4.4} \\

Sitting &
\meanstd{\mathbf{67.2}}{1.9} &
\meanstd{64.8}{5.0} &
\meanstd{56.5}{2.4} &
\meanstd{53.4}{4.2} \\

Lying Down &
\meanstd{64.7}{4.9} &
\meanstd{\mathbf{65.2}}{4.2} &
\meanstd{57.4}{4.8} &
\meanstd{53.1}{5.6} \\

Standing/Walking &
\meanstd{\mathbf{62.5}}{3.5} &
\meanstd{60.3}{2.0} &
\meanstd{50.9}{4.8} &
\meanstd{51.2}{3.9} \\
\hline
Overall &
\meanstd{\mathbf{70.2}}{1.2} &
\meanstd{67.7}{1.5} &
\meanstd{63.3}{0.7} &
\meanstd{56.3}{1.8} \\
\hline
\end{tabular}
\end{table}


Among individual activities, dynamic transitional actions yield systematically higher identification accuracy than continuous postural states, averaging $74.2\%$ vs. $64.8\%$ for DS-SDPNet. Sit-to-Lie achieves the peak performance across all backbones, up to $80.8\pm4.2\%$ with DS-SDPNet. This advantage stems from large, whole-body kinematic trajectories that provide rich individualized motion signatures. In contrast, stationary states, including Sitting at $67.2\%$ and Lying Down at $64.7\%$, rely primarily on body-geometry configurations and subtle micro-movements, resulting in lower separability. Standing/Walking yields the lowest accuracy of $62.5\pm3.5\%$, which we attribute to the temporal heterogeneity of interleaving stops, turns, and intermittent strides within short clips.

\subsection{HAR Router Performance}

We next evaluate whether the HAR router can accurately predict activity labels to guide expert selection when the activity is unknown at inference time. We implement a leave-one-subject-out cross-validation protocol, with classification results across seven ADLs reported in Table~\ref{tab:per_activity_har_accuracy}.   Our proposed PointNet-LSTM router achieves an overall mean accuracy of $95.7\pm4.0\%$, outperforming point cloud activity recognition baselines DVCNN~\cite{yu2022noninvasive} ($90.8\pm5.6\%$) and m-Activity~\cite{wang2021m} ($86.9\pm6.1\%$). It achieves the top accuracy across all seven activities, and also exhibits tighter cross-subject variance than the baselines. These results confirm that the router provides robust and reliable activity routing across unseen subjects, effectively minimizing error propagation to downstream identity experts.

\begin{table}[h]
\centering
\caption{Per-activity HAR accuracy on mm-ADL (\%).}
\label{tab:per_activity_har_accuracy}
\renewcommand{\arraystretch}{1.08}
\setlength{\tabcolsep}{4.0pt}
\begin{tabular}{lccc}
\hline
Activity & PointNet-LSTM & DVCNN & m-Activity \\
\hline
Sit-to-Lie &
\meanstd{\mathbf{94.9}}{7.0} &
\meanstd{88.6}{12.5} &
\meanstd{82.9}{14.1} \\

Lie-to-Sit &
\meanstd{\mathbf{96.9}}{4.9} &
\meanstd{89.8}{10.2} &
\meanstd{86.4}{11.9} \\

Sit-to-Stand &
\meanstd{\mathbf{95.6}}{5.7} &
\meanstd{95.3}{5.7} &
\meanstd{93.1}{9.4} \\

Stand-to-Sit &
\meanstd{\mathbf{94.0}}{5.2} &
\meanstd{89.1}{8.1} &
\meanstd{87.5}{6.4} \\

Sitting &
\meanstd{\mathbf{94.5}}{3.9} &
\meanstd{84.4}{7.6} &
\meanstd{82.9}{10.3} \\

Lying Down &
\meanstd{\mathbf{95.8}}{3.9} &
\meanstd{92.9}{6.5} &
\meanstd{89.5}{8.6} \\

Standing/Walking &
\meanstd{\mathbf{98.4}}{1.9} &
\meanstd{95.5}{3.9} &
\meanstd{86.2}{12.0} \\
\hline
Overall &
\meanstd{\mathbf{95.7}}{4.0} &
\meanstd{90.8}{5.6} &
\meanstd{86.9}{6.1} \\
\hline
\end{tabular}
\end{table}

\subsection{Comparison of Activity--Identity Learning Strategies}

Using the DS-SDPNet identification backbone and PointNet-LSTM HAR router, we evaluate three paradigms for incorporating activity information into closed-set identification when activity labels are unavailable during inference: monolithic learning, joint multi-task learning (MTL) and two-stage activity-conditioned routing.


Table~\ref{tab:routing_study} reports identification accuracy across all strategies alongside the theoretical oracle upper bound. Monolithic learning achieves an accuracy of $62.1\pm1.2\%$. Introducing activity supervision via joint MTL increases performance to $64.3\pm1.5\%$ ($+2.2\text{ pp}$), confirming that activity cues provide beneficial regularization for identity representations. However, because both monolithic and joint MTL architectures enforce a shared feature space across all ADLs, the shared representation must accommodate substantial motion and postural shifts across activities, which may constrain its capacity for fine-grained identity discrimination. In contrast, the two-stage hard-routing framework achieves $68.0\pm1.1\%$, improving upon the monolithic baseline by $+5.9\text{ pp}$ and joint MTL by $+3.7\text{ pp}$, while falling just $2.2\text{ pp}$ below the oracle upper bound of $70.2\%$. These results suggest that decoupling activity routing from identity modeling allows individual experts to focus on identity cues specific to each motion pattern, rather than forcing a single model to reconcile wide postural and kinematic variations across all ADLs.

\begin{table}[h]
\centering
\caption{Closed-set ID under different activity--identity learning strategies (\%).}
\label{tab:routing_study}
\renewcommand{\arraystretch}{1.1}
\setlength{\tabcolsep}{7pt}
\begin{tabular}{lc}\hline
Method & ID accuracy \\ \hline
Monolithic & \meanstd{62.1}{1.2} \\
Joint multi-task learning & \meanstd{64.3}{1.5} \\
Two-stage activity-conditioned & \meanstd{68.0}{1.1} \\ \hline
Oracle activity selection & \meanstd{70.2}{1.2} \\ \hline
\end{tabular}
\end{table}

\begin{table*}[!t]
\centering
\caption{Subject-disjoint ReID: primary strategy configurations and controlled retrieval comparisons (\%).}
\label{tab:reid_gallery}
\renewcommand{\arraystretch}{1.12}
\setlength{\tabcolsep}{9pt}
\begin{tabular}{llcc}
\hline
Retrieval protocol & Identity representation & mAP & Rank-1 \\
\hline

\multicolumn{4}{l}{\textit{Primary strategy configurations}} \\
Global gallery &
Monolithic &
$57.2\pm5.1$ & $59.1\pm6.9$ \\
Global gallery &
Joint multi-task learning &
$56.4\pm4.1$ & $84.6\pm9.0$ \\
HAR-partitioned gallery &
Activity-specific experts (two-stage) &
$75.4\pm13.0$ & $82.1\pm12.3$ \\
\hline

\multicolumn{4}{l}{\textit{Gallery-matched controls}} \\
HAR-partitioned gallery &
Monolithic &
$61.5\pm7.2$ & $62.9\pm15.9$ \\
HAR-partitioned gallery &
Joint multi-task learning &
$68.2\pm9.9$ & $82.8\pm11.8$ \\
\hline

\multicolumn{4}{l}{\textit{Oracle activity reference}} \\
Oracle-partitioned gallery &
Activity-specific experts (two-stage) &
$76.2\pm12.8$ & $84.1\pm14.0$ \\
\hline
\end{tabular}

\end{table*}

\section{Subject-Disjoint ReID Results}\label{sec:results_reid}
\subsection{Native Retrieval Performance}

We first examine whether the learned identity representations generalize to subjects excluded from training. Following the subject-disjoint protocol detailed in Section~\ref{sec:experiments}, two held-out subjects are evaluated per fold without test-time model updating. In their native configurations, the monolithic and joint multi-task models operate over a global gallery containing enrolled clips from all activities, whereas the two-stage model uses the HAR-predicted activity to search an activity-partitioned gallery. The top block of Table~\ref{tab:reid_gallery} summarizes these primary strategy configurations.


Under global-gallery retrieval, the monolithic baseline achieves $57.2\%$ mAP and $59.1\%$ Rank-1. Joint multi-task learning dramatically improves first-match retrieval to $84.6\%$ Rank-1 (+25.5 pp), though its overall ranking quality remains constrained at $56.4\%$ mAP. In contrast, the two-stage activity-conditioned pipeline delivers strong first-match accuracy of $82.1\%$ Rank-1 alongside a substantially higher mean mAP of $75.4\%$ (+18.2 pp over monolithic). Because native pipelines couple distinct representation schemes with different gallery search spaces (global vs. partitioned), these initial comparisons reflect end-to-end operational performance. To isolate the representation gains from the gallery partitioning effect, we evaluate gallery-matched controls in the following subsection. 

\subsection{Influence of Gallery Partitioning}


The performance advantage of the two-stage framework arises from two factors: restricting retrieval to a smaller same-activity candidate pool, and learning specialized activity-dependent embeddings. To isolate the impact of gallery pruning, the middle block of Table~\ref{tab:reid_gallery} evaluates the monolithic and joint MTL representations when restricted to the identical HAR-partitioned gallery used by the two-stage pipeline.

Restricting the search space to the predicted activity subset improves retrieval quality across both shared models. For the monolithic baseline, mAP rises from $57.2\%$ to $61.5\%$ and Rank-1 increases from $59.1\%$ to $62.9\%$. For joint MTL, gallery partitioning yields a substantial boost in mAP, from $56.4\%$ to $68.2\%$, while preserving high first-match retrieval of $82.8\%$ vs. $84.6\%$ Rank-1. This indicates that joint MTL already encodes effective within-activity identity features, but its overall ranking quality was previously suppressed by cross-activity gallery distractors.  



\subsection{Identity Representations Under Activity-Partitioned Retrieval}


With the gallery search space held constant under HAR partitioning, we directly evaluate the discriminative power of the three underlying identity representations. Under this controlled setting, monolithic learning achieves $61.5\%$ mAP and $62.9\%$ Rank-1, while joint multi-task learning reaches $68.2\%$ mAP and $82.8\%$ Rank-1. The proposed activity-specific expert bank attains $75.4\%$ mAP and $82.1\%$ Rank-1. 

While joint MTL provides competitive first-match performance ($82.8\%$ vs. $82.1\%$ Rank-1), the activity-specific expert bank delivers decisively superior overall ranking quality, surpassing joint MTL by $+7.2\text{ pp}$ and monolithic learning by $+13.9\text{ pp}$ in mean mAP. This demonstrates that activity-conditioned specialization significantly increases precision across the entire retrieved list, confirming that the two-stage framework's gains are fundamentally driven by representation quality rather than search-space reduction alone.  

Furthermore, as shown by the oracle reference in Table~\ref{tab:reid_gallery}, eliminating routing errors pushes the activity-specific representation to $76.2\%$ mAP and $84.1\%$ Rank-1. This indicates that activity-specific specialization inherently supports top-tier performance in both metrics, with the slight gap in practical Rank-1 between $82.1\%$ and $84.1\%$ stemming primarily from router classification errors rather than limits of the identity representation itself.   



\subsection{Learned Versus Oracle Activity Routing}

To evaluate the penalty incurred by automated activity estimation, we compare the two-stage pipeline under learned PointNet-LSTM routing against the theoretical oracle upper bound (Table~\ref{tab:reid_gallery}). Replacing oracle routing ($76.2\%$ mAP, $84.1\%$ Rank-1) with learned routing ($75.4\%$ mAP, $82.1\%$ Rank-1) results in marginal decreases of only $0.8\text{ pp}$ and $2.0\text{ pp}$, respectively. This demonstrates that the router’s high classification fidelity, with $95.7\%$ accuracy, virtually eliminates error propagation to downstream ReID experts.

We further analyze whether misrouted gallery clips (activity contaminants) degrade retrieval by evaluating an offline control where such false insertions are filtered out using ground-truth activity labels. Removing these contaminants alters retrieval performance by merely $+0.4\text{ pp}$ mAP and $-0.6\text{ pp}$ Rank-1, confirming that gallery contamination has negligible practical impact. Because the activity-specific experts are optimized for their respective kinematic manifolds, out-of-activity clips typically map to distant embedding regions and rarely displace genuine, in-activity matches in the top ranks.

\section{Backbone and Representation Analysis}\label{sec:backbone_analysis}

\subsection{Identity Backbone Comparison on mm-ADL}
We reproduce three mmWave point-cloud identity baselines under the same input and folds. SRPNet~\cite{cheng2021person} extracts spatio-temporal features for radar point-cloud ReID. HDNet~\cite{huang2023hdnet} uses hierarchical motion modeling with a point-flow descriptor and dynamic frame sampling. DSFE+LGTE~\cite{xue2023fine} combines dual-stream spatial feature extraction with local-global temporal encoding. Although these models were developed mainly for gait, they provide closely related point-cloud identity baselines for evaluating ADL identification.

DS-SDPNet achieves the highest overall mean closed-set ID accuracy among the evaluated backbones on mm-ADL, 70.2\%, compared with 67.7\% for HDNet, as reported in Table~\ref{tab:backbone_oracle_routing}. It also has the highest mean ID accuracy on six of the seven activities in Table~\ref{tab:per_activity_accuracy_comparison}. Under the oracle routing condition in Table~\ref{tab:backbone_oracle_routing}, the ReID evaluation yields the same relative backbone ranking as closed-set ID, with DS-SDPNet consistently outperforming all baselines across both mAP and Rank-1 metrics.

\begin{table}[h]
    \centering
    \caption{Backbone comparison under oracle routing (\%).}
    \label{tab:backbone_oracle_routing}
    \renewcommand{\arraystretch}{1.10}
    \setlength{\tabcolsep}{2.0pt}
    \begin{tabular}{lccccc}
        \hline
        Backbone &
        \shortstack{ID\\mm-ADL} &
        \shortstack{ID\\mmGait-10} &
        \shortstack{ReID\\mAP} &
        \shortstack{ReID\\Rank-1} &
        FLOPs \\
        \hline
        SRPNet &
        \meanstd{56.3}{1.8} &
        \meanstd{35.3}{2.7} &
        \meanstd{58.5}{7.4} &
        \meanstd{61.1}{17.1} &
        1.19G \\

        HDNet &
        \meanstd{67.7}{1.5} &
        \meanstd{80.8}{2.6} &
        \meanstd{75.5}{13.8} &
        \meanstd{81.9}{15.5} &
        3.49G \\

        DSFE+LGTE &
        \meanstd{63.3}{0.7} &
        \meanstd{65.1}{5.2} &
        \meanstd{69.2}{11.9} &
        \meanstd{79.9}{14.2} &
        0.85G \\

        DS-SDPNet &
        \meanstd{\mathbf{70.2}}{1.2} &
        \meanstd{\mathbf{81.1}}{0.5} &
        \meanstd{\mathbf{76.2}}{12.8} &
        \meanstd{\mathbf{84.1}}{14.0} &
        1.24G \\
        \hline
    \end{tabular}
\end{table}

\subsection{Cross-Dataset Evaluation on mmGait}

We reproduce the evaluated identity backbones on the single-person portion of mmGait using the same 20-frame window and five-fold protocol. Although DS-SDPNet achieves a mean ID accuracy numerically close to HDNet ($81.1\%$ vs. $80.8\%$), it demonstrates markedly superior cross-fold consistency, reducing standard deviation by over a factor of five ($0.5\%$ vs. $2.6\%$). Furthermore, DS-SDPNet delivers this more stable and slightly higher recognition rate with substantially greater computational efficiency, requiring only $1.24\text{ G}$ FLOPs compared to $3.49\text{ G}$ for HDNet.

\subsection{Component Ablation}

Table~\ref{tab:component_ablation} summarizes the component ablations. For HAR, spatial modeling via PointNet alone achieves 92.3\% accuracy, outperforming the temporal-only LSTM baseline of 85.6\%. Integrating the LSTM with PointNet yields the highest accuracy of 95.7\% while substantially lowering cross-fold variance. 



For ID, the dynamic-only variant achieves $69.6\%$ mean accuracy across activities, compared with $67.1\%$ for the static-only stream. While the overall mean gain of DS-SDPNet over the dynamic stream appears modest, the combined architecture is critical for stabilizing harder classes: activity-level breakdown shows notable gains in Standing/Walking ($+4.3$ pp) and Sitting ($+3.2$ pp). Moreover, as reflected by the reduced standard deviation in and an improved worst-case activity accuracy (from $58.1\%$ to $62.5\%$), the dual-stream design achieves considerably more uniform identification across all ADLs.

\begin{table}[h]
\centering
\caption{HAR-router and DS-SDPNet component ablations (accuracy in \%).}
\label{tab:component_ablation}
\renewcommand{\arraystretch}{1.08}
\setlength{\tabcolsep}{3.5pt}

\begin{threeparttable}
\begin{tabular}{llcc}
\hline
Component & Configuration & Accuracy & FLOPs \\
\hline
\multirow{3}{*}{HAR router}
& LSTM-only &
\meanstd{85.6}{6.0} &
13.68M \\

& PointNet-only &
\meanstd{92.3}{5.4} &
752.79M \\

& PointNet-LSTM &
\meanstd{\mathbf{95.7}}{4.0} &
776.66M \\
\hline
\multirow{3}{*}{ID expert}
& Static stream &
\meanstd{67.1}{6.6} &
480.49M \\

& Dynamic stream &
\meanstd{69.6}{7.3} &
763.78M \\

& DS-SDPNet &
\meanstd{\mathbf{70.2}}{5.8} &
1.24G \\
\hline
\end{tabular}

\begin{tablenotes}[flushleft]
\footnotesize
\item[] \textit{Note:} For ID experts, values denote
mean $\pm$ standard deviation across seven activities.
\end{tablenotes}

\end{threeparttable}
\end{table}

\begin{figure*}[!t]
\centering
\includegraphics[width=\textwidth]{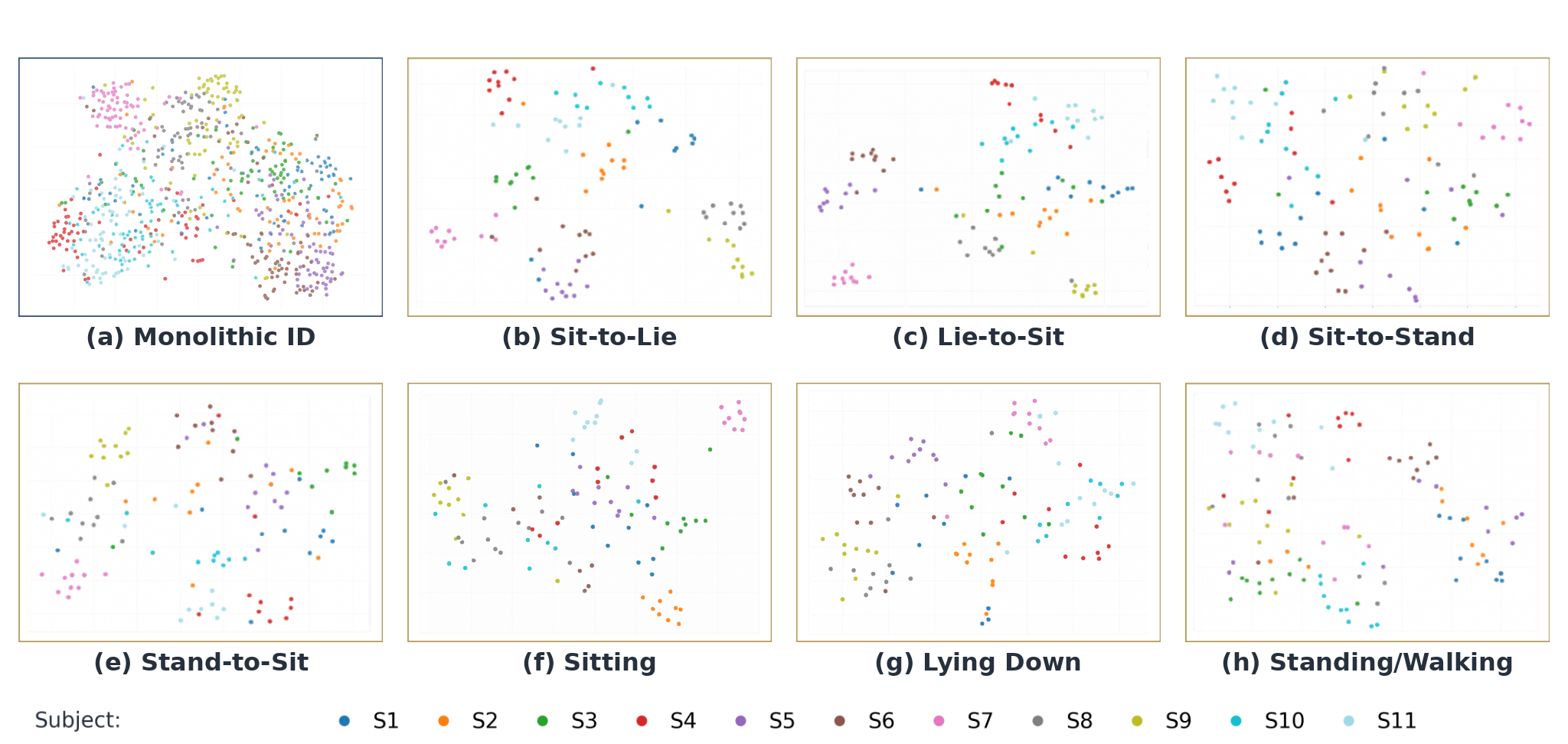}
\caption{Identity-colored t-SNE visualizations of penultimate-layer features from held-out ID samples. (a) Monolithic DS-SDPNet trained across all activities. (b)--(h) Activity-specific DS-SDPNet experts, ordered by ID accuracy. Colors denote subject identities (S1--S11).}
\label{fig:id_comparison}
\end{figure*}

\subsection{Identity Feature Visualization}\label{subsec:feature_visualization}


To evaluate how activity conditioning impacts identity discrimination, we visualize the penultimate-layer feature spaces using t-SNE on held-out test subjects, as shown in Fig.~\ref{fig:id_comparison}. In the monolithic representation (Fig.~\ref{fig:id_comparison}(a)), embeddings from different subjects are heavily interspersed, demonstrating that inter-activity variance confounds subject identity when a single model handles all ADLs. In contrast, the activity-conditioned experts (Fig.~\ref{fig:id_comparison}(b)–(h)) yield substantially tighter intra-subject clustering and clearer inter-subject separation. This identity disentanglement is particularly pronounced in transitional activities such as Sit-to-Lie and Lie-to-Sit (Fig.~\ref{fig:id_comparison} (b) and (c)), where subjects form distinct, well-isolated clusters. While static states such as Sitting and Lying Down (Fig.~\ref{fig:id_comparison} (f) and (g)) exhibit comparatively broader cluster spread due to reduced motion cues, they nonetheless retain better local identity coherence than the monolithic baseline. These qualitative patterns confirm that conditioning on activity simplifies the identity manifold and underpins the quantitative retrieval gains.

\section{Limitations and Future Work}\label{sec:limitations}

While this work demonstrates the feasibility of mmWave point-cloud identification and re-identification across daily activities, the evaluation is under controlled conditions with a fixed radar viewpoint and scripted execution protocols. While this setup isolates subject morphology and motion kinematics from external confounders, practical deployments introduce varying aspect angles, dynamic multipath clutter, distance-dependent point density variations, and unconstrained intra-activity styles. Investigating viewpoint-invariant representations and multi-radar configurations will be essential to address these factors.

Furthermore, the discrete seven-activity ethogram models daily behaviors coarsely compared to continuous, unsegmented domestic activities. Future extensions should develop continuous temporal segmentation and composite action modeling. Finally, due to the high annotation overhead of 3D point clouds, the mm-ADL dataset is limited to 11 subjects and 50 clips per activity. Validating the framework on larger cohorts with broader gallery distractor pools, sparser enrollment samples per subject, integrating spatiotemporal trajectory tracking for identity persistence, and supporting explicit open-set unknown rejection remain key avenues for future work.

\section{Conclusion}\label{sec:conclusion}

In this study, we demonstrated the feasibility of person identification and re-identification from millimeter-wave radar point clouds across diverse activities of daily living. To address the entanglement of activity and identity cues, we systematically evaluated three activity–identity learning paradigms under both closed-set identification and subject-disjoint retrieval settings, characterizing their distinct architectural trade-offs. In parallel, our proposed DS-SDPNet enhanced biometric feature extraction by effectively coupling spatial distribution with temporal motion dynamics. Together, these findings show that discriminative biometric signatures can be reliably extracted beyond gait, enabling privacy-preserving mmWave sensing to potentially extend beyond designated passages into broader, unconstrained indoor environments.


\section*{Acknowledgment}
We are grateful to NSERC for partially funding this project through the I2I grant titled ``AI-based continuous, non-invasive and unobtrusive monitoring of cardiac patients.''

\bibliography{reference}
\end{document}